# Когнитивная архитектура для систем поддержки принятия решений, основанная на принципах функционирования мозга

Антон Колонин[a], Андрей Курпатов[b], Артем Молчанов[b], Геннадий Аверьянов[b]

[a]Novosibirsk State University, Pirogova 2, Novosibirsk, 630090, Russia
[b]Sberbank of Russia, Neuroscience Lab, Vavilova 19, Moscow, 117312, Russia

В данной статье мы описываем когнитивную архитектуру, предполагаемую для решения широко спектра систем поддержки принятия решений (СППР), на основе выявленных ранее пяти принципов мозговой активности, с реализацией их в трех подсистемах - логико-вероятностного вывода, вероятностных формальных понятий и теории функциональных систем. Построение архитектуры предполагает реализацию задачного подхода, позволяющего определять целевые функции прикладных приложений как задачи, сформулированные в терминах соответствующей задаче операционной среды, выраженной в прикладной онтологии. Мы приводим базовую онтологию для ряда практических приложений и основанные на ней предметные онтологии, описываем предлагаемую архитектуру и приводим возможные примеры исполнения указанных приложений в данной архитектуре.

## 1. Концепция когнитивного ядра - 5 принципов, 1 подход, 3 составляющих

Предполагается построение универсальной системы поддержки принятия решений (СППР, или Decision Support System - DSS) на основе 5 принципов функционирования мозга ("Brain Programming Principles" или BPP) [1,2] в рамках задачного подхода (ЗП) [2,3,4] с реализацией их в когнитивной архитектуре, определяемой в для произвольно заданной операционной среде [5], действующей на основе логико-вероятностного вывода (ЛВВ) [6], вероятностных формальных понятий (ВФП) [7] и теории функциональных систем (ТФС) [2,3,4].

### 1.1. 5 принципов функционирования мозга

Пять принципов функционирования мозга, сформулированные в [1], предполагают одновременное задействование большей части из них в большинстве когнитивных процессов. То есть, при принятии решения в любой новой ситуации: 1) действие принципа "генерации сложности" приводит к порождению ряда гипотез о расширенном контексте ситуации по ограниченному числу стимулов или восприятий в различных модальностях; 2) принцип "отношения" позволяет соотнести различные стимулы и гипотезы друг с другом, обогащая одни и исключая другие; 3) принцип "аппроксимации до сущности" позволяет приводить опосредованные стимулами частные проявления к понятиям и явлениям, известным из накопленного прошлого опыта; 4) действие принципа "локальности-распределенности" обеспечивает "сборку" полноценных интегрированных контекстов на основании множества частных стимулов и понятий одной или различных модальностях; 5) принцип "тяжести" определяет доминантный контекст, соответствующий наиболее вероятной гипотезе - например, как показано на Рис.1.

# 5 принципов BPP в принятии решений

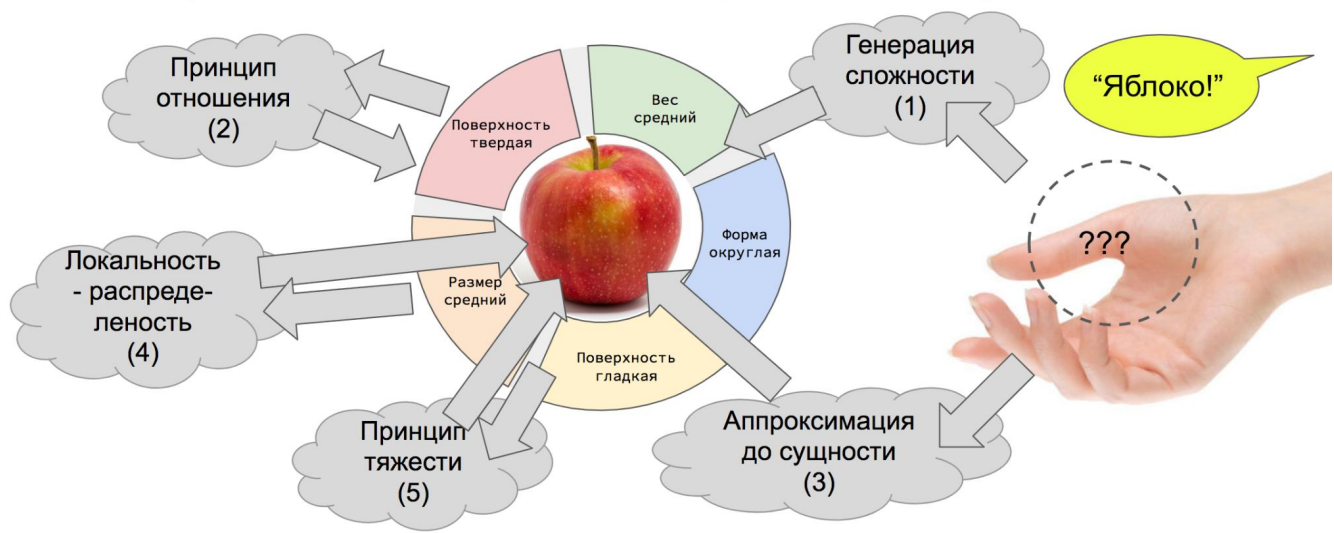

**Рис. 1.** Применение пяти принципов функционирования мозга деятельности (Brain Principles Programming, BPP) к решению когнитивной задачи о распознавании невидимого объекта по его ощущениям.

Например, как показано на Рис 1. Ощущение круглого предмета в одной руке испытуемого, при закрытых глазах, позволяет сгенерировать ряд гипотез, наделяя неизвестный пока объект, в силу принципа "генерации сложности", различными не фиксируемыми пока свойствами, связанными с этим рядом гипотез (оранжевый апельсин, белый бильярдный шар, красное яблоко, зеленое яблоко). Принцип "отношения" позволяет соотнести эти гипотезы с ощущаемым свойствами, делая заключения, что шершавость гипотетического апельсина не соответствует твердости ощущаемого предмета, как и тяжесть гипотетического бильярдного шара не соответствует его легкости, а гладкость и легкость яблок совпадают с реальными ощущениями. Принцип "локальности-распределенности" позволяет собрать в единый образ ощущения гладкой поверхности - со всех пальцев и ладони, небольшого веса - с предплечья и локтя, яблочно запаха - от органов обоняния, а также информацию о нахождении в продуктовом магазине, а не в бильярдном клубе - из краткосрочной памяти. За счет принципа "тяжести", гипотеза о "яблоке" становится доминирующей, подавля гипотезы об "апельсине" и "бильярдном шаре". Наконец, принцип "аппроксимации до сущности" кристаллизует образ конкретно осязаемого яблока, во всех его проявлениях, до абстрактного стереотипного яблока в мыслях испытуемого, позволяя ему уверенно дать правильный ответ "это - яблоко!".

### 1.2. Задачный подход

Задачный подход (ЗП) к проблеме искусственного интеллекта, сформулированный в [2,3,4] на основании Теории Функциональных Систем (ТФС) П.К.Анохина, предполагает возможность решения широкого класса когнитивных задач. Необходимым условием для этого является их формализация в соответствующем операционном пространстве в терминах онтологии конкретной предметной области. Указанная формализация должна обеспечивать возможность описания в этих терминах как самой решаемой задачи, так и образ достижимого результата с учетом критериев его достижения. Последнее может быть сделано введением функции "успешности" (решения задачи) или "полезности" (субъекта, решающего данную задачу либо когнитивной модели, им для этого используемой) определенной в формальных терминах соответствующей операционной среды.

Наряду с этим, как исходные условия решения задачи (включая фиксированные исходные и конечные состояния) так и шаги её решения (бизнес-процессы, исполняемые скрипты и сценарии, решающие правила и правила принятия решений) также должны выразиться в терминах той же самой предметной онтологии.
В работе [5] показана принципиальная возможность решения задач экспериментального, или основанного на собственном опыте взаимодействия с операционной средой, обучения как в рамках различных онтологий одной и той-же предметной области, так и на на основе различных алгоритмов обучения. В этой работе мы предполагаем возможность решения широкого круга задач на основе когнитивной архитектуры, определяемой для произвольной операционной среды, заданной соответствующей онтологической модели. Применительно к классам задач, связанному как с управлением бизнес-процессами, так и с психотерапевтической практикой [9], мы предлагаем использование базовой процесс-ориентированной (или "деятельностной") онтологии [8] в качестве основы для описания конкретных прикладных онтологий. Предложенная в указанной [8] работе "деятельностная онтология", предлагает оперирование, в терминах работы [1], так называемыми "интеллектуальными объектами" и "интеллектуальными функциями". Основными классами интеллектуальных объектов рассматриваются "инварианты" и "экземпляры" ("прецеденты") - первые являются устойчивыми, выраженными в пространстве и времени, представлениями вторых, в то время как вторые являются "доказательной базой" для идентификации первых. В свою очередь, "интеллектуальные функции" представлены несколькими классами когнитивных операций, осуществляемых над "интеллектуальными объектами", как описано ниже.

### 1.3. Математические методы моделирования когнитивной деятельности (ЛВВ, ВФП, ТФП)

Реализация описанного выше задачного подхода на основе перечисленных ранее 5 принципов предполагается на основе трех методов - логико-вероятностного вывода [6], вероятностных формальных понятий [7] и теории функциональных систем [2,3,4], рассматриваемых ниже.

#### 1.3.1. Логико-вероятностный вывод

Метод логико-вероятностного вывода (ЛВВ) [6,10], наряду с системой неаксиоматического логического вывода (Non-axiomatic Reasoning System, NARS) [11] и сетями вероятностной логики (Probabilistic Logic Networks) [12], позволяет решать задачи обучения с учителем, распознавания и предсказания, традиционно связываемые с методами машинного обучения на основе различных регрессионных моделей, включая основанные на искусственных нейронных сетях (ИНС). Предполагаемым достоинством этих методов является принципиальная объяснимость их предсказаний, а также интерпретируемость самих моделей, получаемых в ходе обучения постольку, поскольку в основе их всех лежит семантическое, то есть смысловое представление информации с различием лишь в математических моделях, используемых для расчета вероятностей. Дополнительными достоинством ЛВВ является то, что этот метод, в отличие как других перечисленных методов, а также самих ИНС, позволяет моделировать нейроны человеческого мозга [13], а также совместимость с другими методами (ВФП и ТФС) описанными далее.

В основе метода лежат такие базовые операции расчета вероятностей как "ревизия" (учет накопленного опыта для формирования правил, связывающих причины и следствия), "дедукция" (предсказания на основе полученного опыта, а также вывод новых правил на основе имеющегося опыта), "индукция" (предсказания в условиях неопределенности и генерация гипотез о возможных следствиях на основе имеющихся причин) и "абдукция" (генерация гипотез о возможных причинах в силу имеющихся следствий).

Например, как показано на Рис.2, накопление информации о случаях связи плохого питания с ослаблением иммунной системы дает основания для оценки вероятности правила об обусловленности второго первым в силу операции "ревизии". Далее, наличие правила обусловленности повышенного риска инфицирования ослаблением иммунной системы дает возможность предсказать повышения риска инфицирования в случае плохого питания применяя "дедукцию" к обоим упомянутым правилам, а также создать новое правило на основе этого. Кроме того, наличие установленной связи плохого питания с ростом риска переохлаждения позволяет построить гипотезу о взаимосвязи переохлаждения с ослаблением иммунной системы посредством операции "индукции". Наконец, инфицирование вследствие переохлаждения, при наличии информации о предшествующем снижении иммунитета, также позволяет построить гипотезу о связи последнего с переохлаждением посредством "абдукции".

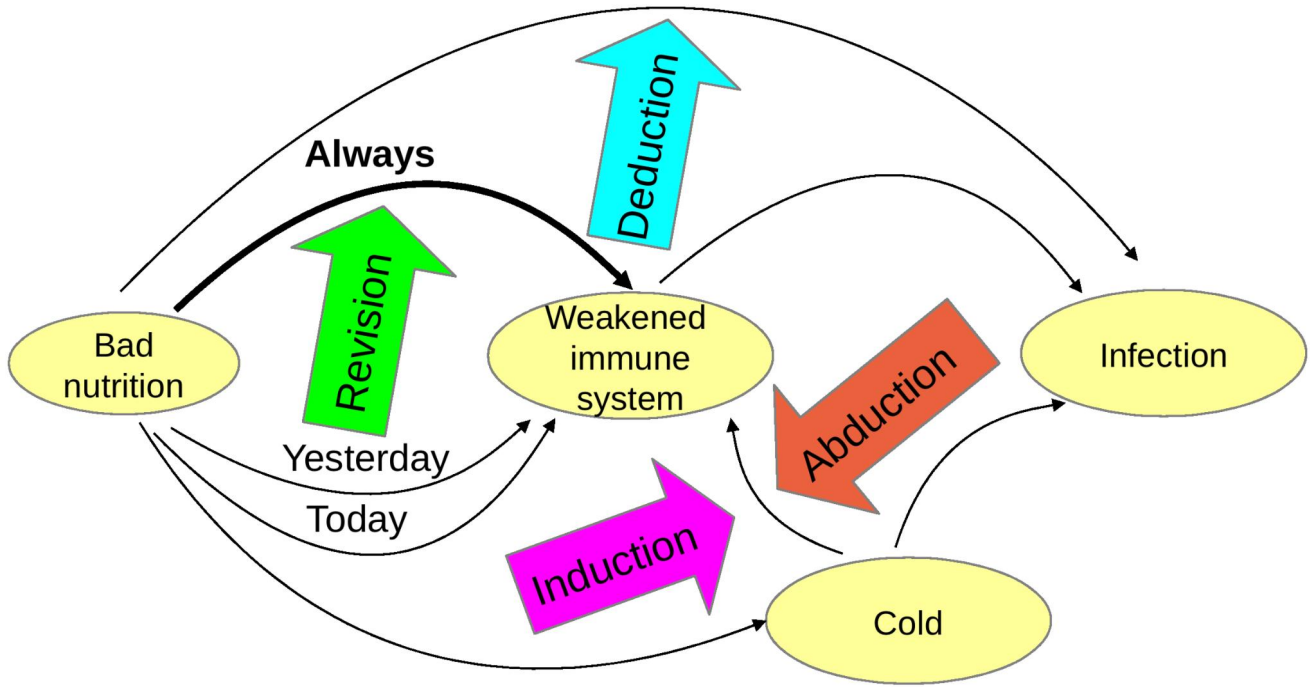

**Рис 2.** Пример использования различных логических операций на примере из предметной области здравоохранения.

#### 1.3.2. Вероятностные Формальные Понятия

Метод вероятностных формальных понятий (ВФП) [7,14] является расширением известного метода анализа формальных понятий (Formal Concept Analysis) на основе методологии ЛВВ описанной выше. Этот метод позволяет выявлять значимые инвариантные сочетания признаков и контекстов как взаимосвязанных причинно-следственных ассоциаций, которые могут быть про интерпретированы как "неподвижные точки" или "аттракторы" в терминах нейросетевой динамики сетей Хопфилда, на основе многократно наблюдаемых или воспринимаемых экземпляров (прецедентов). Указанные инварианты могут рассматриваться как классы объектов или явлений, соответствующих данным прецедентам в пространстве значимых признаков, выявляемых в процессе применения данного метода. Кроме того, метод также потенциально позволяет строить иерархии таких классов как системы понятий, а также иерархий самих признаков как категорий, определяющих данные понятия, и самих значений определяемых для этих категорий.

По своей сути, ВФП предназначен решать широко распространенную в области машинного обучения задачу кластеризации, однако делать это в рамках прозрачной интерпретируемой парадигмы, позволяющей как верифицировать смысловую структуру выявляемых иерархий, так и ассоциировать каждый выявленный инвариант с определяющими его правилами, а также верифицировать сами эти правила.

Например, в контексте задачи распознавания фруктов на Рис.1, ручной перебор и осмотр всех фруктов на прилавке с обработкой данных восприятий методом ВФП позволит выявить такие высокоуровневые инварианты, как фрукты и ягоды, причем фрукты могут включать в себя инварианты апельсинов, яблок и груш, а яблоки - дробиться на инварианты зеленых, желтых и красных яблок. Вместе с тем, значимыми признаками в данной системе понятий окажутся вес, размер, твердость и гладкость поверхности, а также - цвет, причем области определения возможных значений цвета будут разными у различных инвариантов верхнего уровня (например, цитрусовые могут быть только зелеными, красными, оранжевыми или желтыми, но - не синими).

#### 1.3.3. Теория Функциональных Систем

Прикладное развитие метода на основе теории функциональных систем (ТФС) [2,3,4] направлено на решение задачи, являющейся обобщением так называемого "обучения с подкреплением" (Reinforcement Learning, RL) в машинном обучении. Традиционно RL предполагает обучение на основе положительной или отрицательной обратной связи от среды в виде положительного либо отрицательного подкрепления по факту решения тех или иных задач, либо провалу в их решении. Причем большинство распространенных, на сегодняшний день, моделей RL в машинном обучении не являются объяснимыми и, тем более, интерпретируемыми. Вместе с тем, ТФС позволяет расширить класс решаемых задач на случаи, где внешнее подкрепление в явном появляется редко, отложено или вообще отсутствует, и необходимо самостоятельное обучение (self-supervised learning). В ТФС это достигается посредством базового механизма подкрепления системы за достижение ожидаемого образа результата действия, сформированного в начале действия и фиксируемого при её достижении акцептором результата действия.Такая расширенная постановка задачи соответствует скорее "обучению на основе опыта" (Experiential Learning, EL) [5], включающего в себя как самообучение методом проб и ошибок, так и обучение на основе явных или неявных положительных либо отрицательных подкреплений как от окружающей среды, так и учителя.

Например, при самообучения игре в пинг-понг в известной постановке RL интерпретируемой в терминах EL [5], первая стадия обучения может включать в себя формирование способности отличать движущиеся объекты (шарик и ракетка) от неподвижных (методом ВФП), вторая - учиться управлению ракеткой, связывая привод в действие моторных функций с одним из движущихся объектов, третья - отбиванию мяча, связывая изменение его траектории с пространственной конфигурацией ракетки относительно мяча, четвертая - наконец - собственно правилам игры, необходимым для получением итогового подкрепления.

При всем сказанном выше, в рамках предлагаемой нами когнитивной архитектуры, предполагается совместное использование метода ТФС совместно с ЛВВ и ВСП на основе одних и тех-же структур данных (интеллектуальных объектов), выражаемых в одних и тех же терминах, соответствующих предметной области решаемой задачи.

## 2. Унифицированная деятельностная онтология

С целью обеспечения возможности построения линейки приложений на основе одной и той-же когнитивной архитектуры с использованием описанных выше методов, мы предполагаем наличие простой верхней (базовой) онтологии [8] ориентированной на практическую деятельность, позволяющей описывать метаданные для всей предполагаемой линейки прикладных приложений, рассмотрев три из них в качестве примеров ниже.

### 2.1. Верхняя (базовая) онтология

За основу структуры метаданных, следуя стандартам OWL (https://www.w3.org/TR/owl-ref/) и Schema.org (https://schema.org/), мы принимаем базовые понятия Вещь (Thing) и Свойство (Property), но в дальнейша иерархия классов под ними специализируется с учетом нашего понимания интеллектуального объекта (ИО) [1], как универсальной Вещи, обладающей набором Свойств. В этом понимании, следуя концепции деятельностной, либо процесс-ориентированной онтологии [8], ИО могут быть уникальными Экземплярами либо являться абстрактными Инвариантами множества уникальных экземпляров.

Экземпляры могут быть конкретными Значениями (чисел, дат, текстов, любых физических или биологических либо социальных объектов) или же - Прецедентами фиксируемых во времени отдельных Событий, объединяющих этим моменты Совпадений, а также растянутых во времени взаимосвязанных во времени Процессов - объединяющих разновременные События и их Совпадения. Значения могут быть представлены сложными композиционными Сущностями, уникальными в пространстве предметной области (например - известное данное в ощущениях физической яблоко или конкретный Клиент конкретного банка), конкретными атомарными Категориями (например - красный цвет или мужской пол), используемыми для идентификации и описания Сущностей, а также - произвольными уникально фиксируемыми в предметной области Отношениями, объединяющими различные сущности (например, отношения родства между конкретными Клиентами). События же могут подразделяться на Состояния, фиксируемые в заданные моменты времени (например, остаток баланса Счета или возраст Клиента) либо Действия, эти состояния изменяющие (как то финансовая транзакция либо доставка пиццы на день рождения).

Инварианты, в свою очередь, также разделяются на абстрактные Свойства, инвариантные к своим Значениям и вневременные инварианты Событий, Совпадений и Процессов. Инвариантами Событий мы называем явления, инвариантами Процессов - Сценарии, а инвариантами Совпадений могут быть Сцены и Развилки. Сцены, как конъюнктивные множества, объединяют Явления, соответствующие многократно повторяющимся совместно Событиям. Развилки, как дизъюнктивные множества, включают альтернативные Явления, соответствующие различным Событиям, возможные на одном и том же этапе одного и того же

Сценария, зафиксированные в различных процессах (например, явления оплаты наличными или оплаты по карточке в рамках сценария покупки товара в магазине). На более низком уровне инвариантами значений Категорий являются соответствующие Классификаторы (например, Классификатор Свойства Цвет определяет домен с доменными значениями Красный, Желтый и Зеленый), а инвариантами конкретных Сущностей - их абстрактные Образы.

## 2.2. Модель когнитивно-поведенческой терапии

В частном случае решения задачи диагностики в рамках когнитивно-поведенческой терапии (Cognitive Behavioral Therapy, CBT) в контексте [9], с точки зрения ЗП мы рассматриваем задачу постановки Диагноза, являющегося инвариантом, объединяющим значения такого ряда Классификаторов, как Чувство, Эмоция, Состояние, Социальная Ситуация и Когнитивная Ошибка (именуемая "Когнитивное Искажение" [15] в зарубежной психотерапевтической литературе) с соответствующими доменными значениями - Категориями, определенными согласно конкретной психотерапевтической школы или практики. Также, в инвариант могут добавляться характеристики (Психотип и другие) из профиля клиента, если он известен, в рамках модели управления отношения с клиентами, описанной ниже.

Решение указанной задачи предполагается в контексте психотерапевтической сессии, фиксируемой прецедентом Сессия (подкласс Процесса), причем множество Сессий, проведенных по одному (или премерно одному, с точностью до Развилок) Протоколу (подклассу Сценария). Решение задачи осуществляется в диалоге между Клиентом и Психотерапевтом (подклассами Сущности - Экземпляра), осуществляющими Действия в контексте Сессии. Действия Психотерапевта могут быть Диагностическими или Лечебными, а Клиент может реагировать на них посредством Реакций. Все указанные действия могут характеризоваться Речевыми Паттернами (например, паттерны Когнитивных Искажений [15]), связанными с определенными Классификаторами, указанными выше.

В данной модели, в ходе сессионного взаимодействия, психотерапевт сначала осуществляет Диагностические Действия, стимулируя Клиента на Реакции, позволяющие выявить те или иные Категории, совокупность которых по всем используемым классификаторам позволяет выявить инвариант Диагноза в случае успеха сессии, или не выявить ни одного, в случае формального неуспеха. Следует заметить, что формальный неуспех сессии в такой постановке может означать: а) действительной неудачей Психотерапевта выявить реально существующую проблему у клиента; б) отсутствие действительной проблемы у Клиента на самом деле; в) успешное решение проблемы Клиента в ходе самого сессионного Взаимодействия, особенно если часть Действий были Лечебными. В двух последних случаях ЗП позволяет не просто поставить Диагноз (в случае формального успеха), но и позволить приложить максимум усилий на его постановку, обеспечивая снятие всех возможных диагнозов при отсутствии таковых.

## 2.3. Модель управления отношениями с клиентами предприятия

В частном случае формирования предложений Клиенту Предприятия в рамках деятельности по управлению отношений с клиентами (Customer Relationship Management, CRM), задачей, с точки зрения ЗП, является формирование такого Предложения, на которое Клиент отреагирует, с большой вероятностью, положительно - что будет зафиксировано его Подпиской на предлагаемый Продукт. Как Предложение, так и Подписка являются в предлагаемой онтологии Действиями, фиксируемыми, как и в случае CBT, в контексте некоторого краткосрочного Взаимодействия (обращения Клиента в Предприятие или звонок Клиенту из Предприятия). Взаимодействие может быть частью Эпизода взаимодействий Клиент-Предприятие (пользование определенной линейкой сервисов и услуг в определенном офисе продолжительное время), а Эпизод - частью долгосрочной клиентской Истории (включающей все Эпизоды на протяжении времени, когда Клиент был лоялен Предприятию). Сами индивидуальные Клиенты, как и в случае CBT, могут быть определены в многомерном пространстве большого числа Классификаторов, таких как Пол, Поколение, Доход, Социальный Статус, Семейное Положение и многие другие, а также иметь связи с Продуктами Предприятия через отношения Текущая Подписка. При этом устойчивые сочетания Категорий - значений Классификаторов в купе с Текущими Подписками могут образовывать такие Инварианты как Образ Клиента (например - молодой женатый мужчина, пользующийся ипотекой).

Как Клиент с Предприятием, так и Продукт являются подклассами Сущности - Экземпляра, а Продукты, в отношении которых предполагается осуществлять Предложения и имеющие Текущие Подписки, характеризуются Типом Продукта.

Истории, Эпизоды и Взаимодействия являются Процессами - Прецедентами, в которых накапливается историческая информация о Предложениях, сделанных Клиенту, и о Подписках, осуществляемых Клиентом, в контекстах значений его Текущих Подписок, которые могут быть зафиксированы как Состояния Клиента. Агрегация этой информации позволяет выявлять Типовые Истории, Типовые Эпизоды и Типовые Действия как Сценарии - Инварианты, характерные для тех или иных Образов Клиента.

На основании имеющейся выше информации, задача в терминах ЗП может быть специфицирована как формирование, в контексте любого заданного Клиента, Предложений на такие конкретные Продукты и Типы Продукта, чтобы максимизировать вероятность осуществления Подписки на предложенный Продукт, фиксируемой "акцептором результата действия" [2] по оценке результатов доставки Предложения Клиенту.

### 2.4. Модель управления проектами

Частный случай приложения по управлению проектами предусматривает решение задачи по назначению исполнителя для той или иной Задачи того или иного Проекта, где Задачи и Проекты являются Процессами - Прецедентами, причем более атомарные Задачи выполняются в контексте Проектов. Как Проект, так и Задача могут иметь внешнюю оценку "успешности", получаемую извне системы и фиксируемую ей, при этом Проект может иметь текущую агрегированную метрику успешности до своего завершения по совокупности успешности всех входящих в него Задач.

Проекты могут характеризоваться в пространстве таких Классификаторов как Аудитория и Предметная Область, а Задачи, входящие в соответствующие Проекты, определяются в пространстве таких Классификаторов как Приоритет, Серьёзность, Статус и Бюджет, а также ассоциируются с различными Сотрудниками, каждый из которых может занимать определенную Позицию (Директор, Руководитель Отдела, Ведущий Инженер, итд.), а занимаемые Позиции позволяют Сотрудникам выступать в тех или иных Ролях в контекстах как Проектов (Проектная Роль) так и Задач (Задачная Роль).

Задачей (с точки зрения ЗП) в данном контексте является определение для Задачи (в рамках обсуждаемой онтологии) такого Сотрудника (на соответствующей Позицией), присвоением ему соответствующей Задачной Роли, чтобы максимизировать как успешность решения данной задачи, так и всего Проекта, включающего данную задачу, в целом. При этом могут быть учтены исторические Прецеденты о решении похожих Задач в похожих Проектах тем же самым Сотрудником, а также другими Сотрудниками, с учетом итоговых успешностей Задач и Проектов.

## 3. Когнитивная архитектура принятия решений

К когнитивным архитектурам, основанным на онтологических или семантических моделях и вероятностной логике, относятся, прежде всего, Non-Axiomatic Reasoning System (NARS) [11] и OpenCog [16]. В этих системах присутствуют алгоритмические возможности, сходные с ЛВВ в виде собственно не аксиоматической логики [11] и сетей вероятностной логики [12]. Однако в них не предусмотрены предлагаемые нами возможности ЗП, ВФП и ТФС, как это было сделано в наших предыдущих работах [2,5]. В рамках предлагаемой нами архитектуры, общим для всех частных прикладных приложений предполагается использование универсального «когнитивного ядра» (КЯ), исполняющегося на основе частных онтологий соответствующих предметных областей и предоставляющего приложениям верхнего уровня использование возможностей ЛВВ, ТФС и ВФП в рамках ЗП на основе «когнитивной базы данных (знаний)» (КБД).

Сама КБД, как указано на Рис. 3, предназначается для хранения данных и метаданных основанных на описанной выше онтологии верхнего уровня и специализируемых для других предметных областей, предоставляя соответствующий программный интерфейс (API) для когнитивных функций КЯ - ЛВВ, ТФС, ВФП и ЗП.

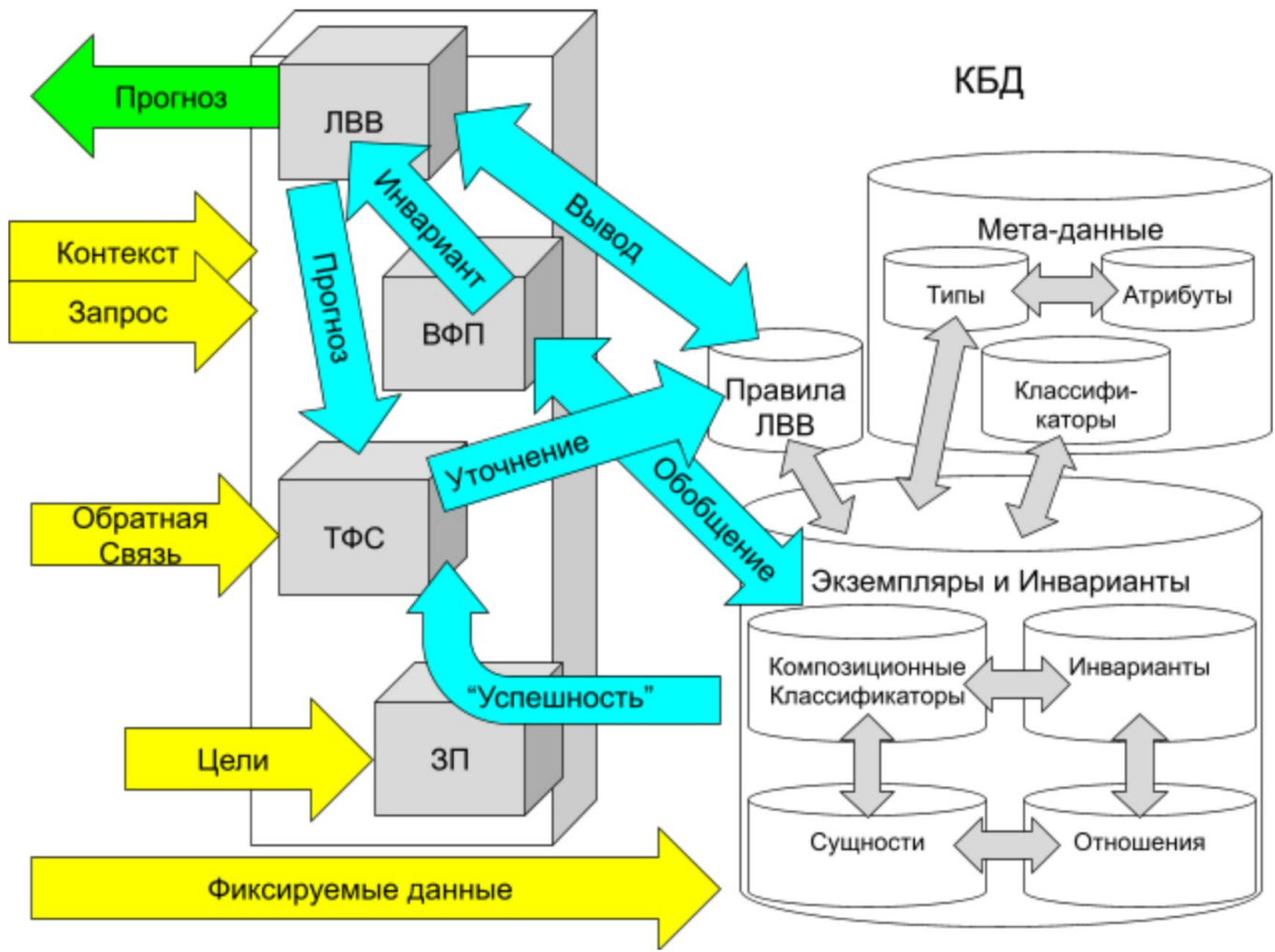

**Рис.3.** Схема когнитивного ядра (КЯ), включающего сервисы ЛВВ, ВФП, ТФС и ЗП на основе КБД.

Примерный состав КЯ, может быть описан как набор описанных далее "сервисов", предоставляющих функции API как друг другу, так и "внешним" приложениями. Таким образом, возможна реализация системы в рамках так называемой "микро-сервисной" архитектуры на основе сервисов ЛВВ, ВФП, ТФС и ЗП на основе единой КБД.

### 3.1. Когнитивная База Данных (КБД)

КБД включает в себя следующие составляющие.
- База Метаданных (БМД) предназначена для хранения онтологических схем экземпляров и инвариантов (посредством описания их типов и атрибутов), а также доменных значений классификаторов. Предполагается полное или частичное кэширование БМД в in-memory графовой БД или специализированной системы в оперативной памяти для быстрого доступа на уровне прикладного слоя для а) получения схем хранения произвольных экземпляров и инвариантов на основе типов и их атрибутов; б) получения текстовых значений объектов данных (в первую очередь - доменных значений классификаторов), а также метаданных - типов и атрибутов для операций с ними.
- База Объектов Данных (БОД) предназначена для хранения, в рамках описанной в БМД выше структуре, как исходной информации об экземплярах и их связях друг с другом, так и производных инвариантов с обеспечением такого функционала, как а) сохранение прецедентов в проекциях реляционной нормализации соответствующей структуре исходных данных; б) Извлечение прецедентов в проекциях реляционной нормализации соответствующей структуре решаемых задач; в) извлечение и сохранение инвариантов в проекциях реляционной нормализации, соответствующей структуре решаемых задач.

- База Правил (БП), для хранения соответствий между конъюнктивно-дизъюнктивными деревьями пред-условий и вероятностными оценками наступления пост-условий, применительно как к прецедентам, так и инвариантам, с обеспечением а) сохранения выявленных правил; извлечения релевантных правил на основе заданных критериев для пред-условий и пост-условий.

Как следует из Рис.3, Типы с Атрибутами БМД описывают структуру Объектов (Экземпляров и Инвариантов) БОД, включая Отношения между ними и Композиционные Классификаторы, которые могут создаваться на основе суперпозиций значений исходных Классификаторов, в том числе на основе выявленных Инвариантов, а Правила БП могут ассоциироваться с Инвариантами, для которых они выявляются.

### 3.2. Сервис ЛВВ

Сервис ЛВВ реализованный согласно [6,10], предусматривает обеспечение следующего функционала: а) формирование правил на основе зафиксированных в БОД Экземпляров (и Прецедентов) и сохранение их в БП; б) построение прогнозов и выдача рекомендаций, на основании предъявленных прецедентов, с использованием как правил, найденных в БП, так и соответствующих инвариантов, найденных в БОД посредством ВФП с возможностью ранжирования прогнозов как по "вероятности" (в терминах [6,10]) так и по "статистической значимости" (критерию Фишера), а также по некоторым образом определенной комбинации того и другого. Построенные прогнозы могут передаваться как внешней прикладной системе для практического использования, так и сервису ТФС, для корректировки содержащейся в виде правил БП модели когнитивного поведения.

Функционирование сервиса может управляться как явными запросами от внешних прикладных приложений, так и фоновыми задания, предназначенными для периодического обновления БП. При этом как явные запросы, так и фоновые задания могут сопровождаться дополнительной контекстной информацией уточняющей запрос.

### 3.3. Сервис ВФП

Сервис ВФП, реализованный согласно [7,14] предназначается для: а) выявления значимых инвариантов на основе данных Экземпляров (и Прецедентов) в БОД; б) сохранения выявленных инвариантов в БОД. Функционирование сервиса может управляться как фоновыми заданиями, предназначенными для периодического обновления инвариантов БОД и связанных с ними Правил БП. При этом фоновые задания могут сопровождаться дополнительной контекстной информацией уточняющей область поиска инвариантов. Также, возможно использование сервиса непосредственными запросами со стороны ЛВВ или прикладных приложений на выявление инвариантов в определенном требуемом прикладном контексте.

### 3.4. Сервис ТФС

Сервис ТФС, реализованный согласно [1,2,3,4] предполагает обеспечение следующего функционала:

а) получение прогнозов ЛВВ и сопоставление их с фактически зафиксированными прецедентами в БОД а также, возможно, явной обратной связью поступающей из внешней среды (одновременно с фиксацией их в БОД) в ответ на выдаваемые прогнозы ЛВВ;

б) корректировка соответствующих Правил в БП, в зависимости от результатов сопоставления прогнозов с их успешностью. В этом смысле, действие ТФС аналогично глобальному подкреплению [5] текущих когнитивных процессов в контексте эпизодической памяти в случае "успешности" текущих результатов.

Функционирование сервиса обеспечивает восприятие как "внешних реакций" от прикладной системы, на основе исполнения ей рекомендаций ЛВВ КЯ так и "внутренних стимулов", порождаемых сервисом ЗП.

### 3.5. Сервис ЗП

Сервис ЗП [2,3,4] предполагается к реализации в качестве специализированного, по отношению к конкретной предметной области, "Оракула", обеспечивающего "внешнее" (по отношению к остальной части архитектуры) "подкрепление" сообразно глобальным "целям". Цели могут считаться заданными "свыше" для конкретной предметной области через определение "функции успешности", вычисляемой сообразно заданным целям и критериям их достижения на основе данных, фиксируемых всей системой в ходе её работы. Оценки

"успешности" "решения задач" ("достижения целей") передаются сервису ТФС в качестве "внутреннего" подкрепления, опосредующего внешние сигналы и реакции на них системы в контексте "функции успешности". Функционирование сервиса может управляться как триггерами на определенные события (как то завершение Задач или Проектов), так и фоновыми задания, предназначенными для периодического подкрепления текущих активностей посредством сервиса ТФС.

# 4. Примеры прикладных приложений

## 4.1. Сессионная диагностика в когнитивно-поведенческой терапии (СВТ)

Рассмотрим пример воображаемого прикладного приложения - чат-бота "Психотерапевт", предназначенного для постановки предварительного диагноза Клиента, обратившегося за помощью к настоящему психотерапевту, как это показано на Рис.4. Приложение может иметь прикладную часть, обеспечивающую следующее: а) собственно функционал чат-бота, связывающий его с мессенджерами типа Telegram и Slack; b) модуль выявления подтверждений ("да", "точно" итд.), отрицаний ("нет", "ни в коем случае", итд.) и речевых паттернов (на основе технологии выявления паттернов [17] и/или либо нейросетевой реализации классификатора текста в пространстве данных паттернов на основе семантической близости [18,19]) для выявления Речевых Паттернов в Реакциях Клиента; с) собственно когнитивное ядро (КЯ) на основе описанной выше когнитивной архитектуры для получения рекомендуемых диагностических Речевых Паттернов для Диагностических Воздействий со стороны Психотерапевта в ответ на получаемые от Клиента Речевые Паттерны, извлекаемые из его Реакций; d) модуль генерации текстовых Диагностических Воздействий на основе рекомендуемых КЯ паттернов (реализуемой с помощью интерпретируемой генерации текстов на основе "грамматики связей" [20] либо пред-обученных языковых моделей на основе "трансформеров" [21], либо других методов [22]); е) модуль текущего клиентского контекста, отслеживающий историю сессии с запоминанием всех диагностических воздействий (с целью подавления повторений) и хранящий текущий диагностический контекст, включающий исключенные и подтвержденные значения классификаторов на текущий момент.

П: Здравствуйте! Как у вас дела?
К: Плохо. Выпнули лузера! (диагностируем #увольнение и #персонализация)
П: Как это случилось? Кто? (уточняем **социальную ситуацию**)
К: Да на работе. Уволили в общем. (диагностируем #увольнение)
П: Вас это пугает или вы обижены на них? (уточняем **эмоцию** - #тревога или #раздражение)
К: Конечно. Мне же ипотеку платить надо. (#тревога)
П: Вас уволили с работы и вы боитесь, что не сможете её оплатить? (#увольнение, #тревога)
К: Да. (подтверждаем диагноз в части **социальную ситуации** и **эмоции**)
П: Что явилось причиной этой ситуации?
К: Мне просто не повезло - от компании ушел инвестор. (диагностируем **когнитивное искажение** - #персонализация)
П: Вы в этом виноваты? (уточняем **когнитивное искажение** - #персонализация)
К: Нет. Просто я чувствую, что все пропало. (меняем диагноз на #катастрофизацию)
П: Вы боитесь, что после увольнения не будет новой работы? (#тревога, #увольнение, #катастрофизация)
К: Точно. Хотя я хороший специалист, на самом деле. (подтверждаем диагноз в части **эмоции**, **социальной ситуации** и **когнитивного искажения** с дальнейшим переходом к корректирующим воздействиям)

**Рис.4.** Пример воображаемого протокола автоматической сессионной диагностики СВТ. П: - реплики Психотерапевнта, К: реплики клиента. В скобках указаны пояснения по действиям проектируемой системы. Хэш-тегом # обозначены доменные значения (Категории) таких Классификаторов как Эмоция, Социальная Ситуация и Когнитивное Искажение.

Как можно видеть из примера на Рис.4, диалог начинается с формальной реплики Психотерапевта. Первая реакция Клиента сразу позволяет поставить частичный предварительный диагноз в пространстве двух Классификаторов - "персонализация" [15] (Когнитивное Искажение) при "увольнении" (Социальная Ситуация). Далее Психотерапевт подтверждает гипотезу в части Социальной Ситуации и осуществляет Воздействие по выяснению Чувства. Ответ Клиента позволяет выявить Чувство "тревоги", после чего

Психотерапевт подтверждает частичный диагноз "тревоги" в связи с "увольнением" и переходит к уточнению Когнитивного Искажения. Следующая реакция Клиента подтверждает предварительный диагноз ("персонализация"), однако попытка явно подтвердить это меняет диагноз на "катастрофизацию". В конце диалога Психотерапевт подтверждает полный окончательный диагноз ("тревога" в связи с "катастрофизацией" "увольнения") и Клиент подтверждает это, тут же давая зацепку в направлении коррекции ситуации посредством последующих Лечебных Воздействий.

### 4.2. Формирование предложений в управлении отношениями с клиентами (CRM)

Примером применения предлагаемой когнитивной архитектуры в задачах системы управления отношениями с клиентами (CRM), описанным в [1], может быть сегментация клиентской аудитории по предсказание качеств Клиентов в перспективе формирования им Предложений Продуктов Предприятия.

**Рис.5.** Исходные данные для сегментации аудитории Клиентов и прогнозирования их качеств в пространстве Классификаторов системы управления отношениями с клиентами (CRM).

В приложении к работе [1] показано, каким образом алгоритм статистической аппроксимации на основе ВФП применим для обнаружения выраженных социальных контекстов, соответствующих Инвариантам отдельным Экземплярам Клиентов, показанным на Рис.5. Всего были исследованы данные о 2784 респондентов (пользователей) социальной сети в пространстве 36 характеристик пользователей, где каждой характеристике соответствует Классификатор такой, как Пол, Место Жительства, Место Рождения, Жизненные Ценности и Отношение к Вредным Привычкам. В результате работы алгоритма статистической аппроксимации на основе ВФП был выявлен 21 Инвариант, представляющий собой выраженные типы представителей пользовательской аудитории. Наиболее характерными из них оказались 8-й и 11-ый инварианты. Например, 8-й инвариант можно охарактеризовать как замужнюю женщину, проживающую в своем родном городе, для которой главное в жизни семья и дети, ценит в людях доброту и честность, в профиле указаны родственники, отношение к алкоголю и курению негативное, небольшое число фото, видео, аудио. В свою очередь, 11-й инвариант можно охарактеризовать как неженатого мужчину, возможно подростка, который также ценит в людях доброту и честность, но главное для него в жизни – это саморазвитие, компромиссно относится к курению и алкоголю.

Полученная, таким образом, сегментация Клиентской аудитории на ряд Образов Клиента, вкупе с фиксацией формирования Прецедентов Предложений тех или иных Продуктов, сделанных отдельным

Клиентам в каждом сегменте, с учетом фиксации последующих, на коротком интервале времени, Прецедентов Подписок Клиентов на те же самые продукты, может позволить ЛВВ строить прогнозы о принятии либо непринятии Предложений на те или иные продукты любым другим Клиентам, с учетом близости Клиента к тому или иному Образу Клиента и с учетом вероятности принятия Предложения пока каждому из возможных Продуктов для данного Образа Клиента.

### 4.3. Принятие решений в управлении проектами

Примером применения данной архитектуры может быть создание персонального ассистента менеджера, позволяющего осуществлять поддержку в принятии решений на основе ЛВВ и ТФС для широкого круга задач, решаемых менеджером, например назначение исполнителей ("задачных" ролей) - на уровне Задач в контексте Проектов. Поддержка в принятии решений может осуществляется гипотетическим сервисом "Ассистент Менеджера" получающего следующие входы и порождающего соответствующие выходы, как описано ниже.

На вход подаются операционные контексты для выдачи рекомендаций, которые автоматически считывается драйверами/плагинами имеющихся корпоративных приложений (messengers, task trackers), используемых менеджерами в повседневной работе и накапливается в форме прецедентов в БОД в качестве непрерывного "лога событий", структурированного в рамках приведенной выше онтологии, включающей Сотрудников, Позиции, Задачные и Проектные Роли а также сами Задачи и Проекты и все связанные с этим Классификаторы. Кроме этого, операционный контекст может вручную вводится самим менеджером или его ассистентом - человеком.

На выходе, с учетом заданного контекста, в виде Задачи и её Статуса, а также - списков возможных исполнителей из числа Сотрудников и их Позиций, функции ЛВВ и ВФП формируют рекомендации исполнителей для конкретной Задачи с тем, чтобы пользователь-менеджер самостоятельно принял нужное решение выбором из списка. Также, возможно автоматическое принятие решений по назначению исполнителя - в это случае, из списка предоставленных ЛВВ опций автоматически будет выбираться позиция, у которых значения "вероятности" и "статистической значимости" превышает "порог уверенности" ("порог доверия"), заданный в качестве либо глобальной настройки системы, либо параметра индивидуальной настройки каждого конкретного пользователя-менеджера.

Принятые, таким образом, решения фиксируются в БОД качестве Прецедентов в контексте текущих состояний Проектов и Задач и могут быть позднее оценены в рамках ТФС с точки зрения ЗП на предмет успешности или неуспешности активностей, с отражением результатов оценки корректировкой используемых ЛВВ моделей для последующих применений.

## 5. Выводы и рекомендации

Мы показали, каким образом пять принципов мозговой деятельности (ВРР) могут быть реализованы в рамках задачного подхода (ЗП) в когнитивной архитектуре, построенной на основе методов логико-вероятностного вывода (ЛВВ), вероятностных формальных понятий (ВФП) и теории функциональных систем (ТФС), а также продемонстрировали принципиальную применимость предлагаемой архитектуры для ряда прикладных задач, формализуемых в терминах соответствующих предметных онтологий на основе базовой деятельностной онтологии. В дальнейших работах мы планируем разработку предложенной архитектуры и апробацию её на реальных задачах в полном объеме.

## Литература